# Local Patch Encoding-Based Method for Single Image Super-Resolution

Yang Zhao, Ronggang Wang, Wei Jia, Jianchao Yang, Wenmin Wang, Wen Gao

*Abstract*—Recent learning-based super-resolution (SR) methods often focus on dictionary learning or network training. In this paper, we discuss in detail a new SR method based on local patch encoding (LPE) instead of traditional dictionary learning. The proposed method consists of a learning stage and a reconstructing stage. In the learning stage, image patches are classified into different classes by means of the proposed LPE, and then a projection matrix is computed for each class by utilizing a simple constraint. In the reconstructing stage, an input LR patch can be simply reconstructed by computing its LPE code and then multiplying the corresponding projection matrix. Furthermore, we discuss the relationship between the proposed method and the anchored neighborhood regression methods; we also analyze the extendibility of the proposed method. The experimental results on several image sets demonstrate the effectiveness of the LPE-based methods.

## 1. Introduction

Single image super-resolution (SISR), also known as image upsampling or image upscaling, is a fundamental technique for various applications in machine vision and image processing, such as digital photographs, image editing, high-definition and ultra–high-definition television, medical image processing, object recognition, and recent face hallucination [34]. The goal of image SR is to recover a high-resolution image (HRI) from a low-resolution image (LRI). How to reconstruct high-quality HRI at low cost is still a challenging task.

One basic type of SR method is the interpolation-based algorithm, such as nearest neighbor, bilinear interpolation, bicubic interpolation, and splines. Unfortunately, these methods often produce unnatural artifacts, such as blurring, ringing, and jagged edges. Thus, many interpolation-based methods have been proposed to suppress unnatural artifacts by means of edge prior knowledge [39], different interpolating grids [50], edge sharpening processes [20] , *etc*. These improved methods are able to refine the sharpness of edges but cannot recover high-frequency details.

Another classic type of SR method is the reconstruction-based method, which imposes a similarity constraint between the downsampling of the reconstructed HRI and the original LRI. Early multi-frame reconstruction-based methods fused multiple LRIs of the same scene to recover an HRI. However, the multiple frames were difficult to align and tended to produce extra artifacts. Recently, many single-image reconstruction-based methods have been proposed by means of various image models or constraints,

Y. Zhao and W. Jia* are with the School of Computer and Information, Hefei University of Technology, 193 Tunxi Road, Hefei 230011, China (e-mail: yzhao@hfut.edu.cn; icg.jiawei@gmail.com)
Y. Zhao, R. Wang, W. Wang, and W. Gao are with the School of Electronic and Computer Engineering, Peking University Shenzhen Graduate School, 2199 Lishui Road, Shenzhen 518055, China (e-mail: zhaoyang@pkusz.edu.cn; rgwang@pkusz.edu.cn; wangwm@pkusz.edu.cn; wgao@pku.edu.cn)
J. Yang is with the Snapchat Inc, Venice, CA 90291, USA (e-mail: jianchao.yang@snapchat.com)

such as, gradient constraints [30,33,41], total variation regularizer [21], the approximated heaviside function-based method [3], and de-blurring-based methods [23,27]. However, the performance of these reconstruction-based algorithms degrades rapidly when the magnification factor becomes very large.

To recover missing details, many example-based or learning-based SR methods have been proposed over the years. Generally, the example-based methods aim to learn the missing high-frequency information from the low-resolution (LR)/high-resolution (HR) example pairs. This type of method was first proposed in [8] and has been further developed in recent years. Typical learning-based SR methods have been proposed, such as neighbor embedding-based methods [2,40], sparse representation-based methods [6],11,14,18,19,43,46,49], and local self-exemplar-based methods [7,10,44]. Although these example-based methods can recover fine details, the computational cost of these methods is quite high. Recently, fast and high-performance models have been successfully applied in the image SR scenario, *e.g.*, random forest-based methods [26], efficient anchor neighborhood regression (ANR)-based methods [15,31,32,45,47,48], and deep neural network-based methods [4,5,13,16,17,29,38]. Moreover, there are other effective models for related image reconstruction or recovery scenarios, such as discrete cosine transform-based methods [36,37] and the weighted nuclear norm minimization method [12].

Generally, the target of learning-based SR methods is to establish a mapping function from the LR space to HR space. For example, traditional learning-based SR models often compute the mapping function by $x_i = f(y_i)$, where $\{x\}_{i=1}^{N_S}$ and $\{y\}_{i=1}^{N_S}$ denote the HR patches and the LR patches, respectively. The mapping function $f(\cdot)$ is needed to be calculated for each LR patch $y_i$, and thus the total computation cost is often too high. To avoid patch-by-patch optimization of the reconstruction weights or coefficients, Timofte *et al.* [31,32] inroduced an efficient way to compute the $f(\cdot)$ by means of the dictionary atoms $\{d\}_{i=1}^{N_D}$ and corresponding HR labels, *i.e.*, $x_i = f(d_i)$. Then, the mapping function $f(\cdot)$ of an LR input $y_i$ can be simply estimated by utilizing the pre-computed $f(\cdot)$ of its nearest atom $d_i$. Most recently, convolutional networks are used to fit the mapping function, *i.e.*, $X = f(Y)$, where $X$ and $Y$ denote the HRI and LRI, respectively. These end-to-end networks can directly reconstruct the entire image and also avoid complex patch-by-patch computation.

There are other ways to avoid patch-by-patch calculation of $f(\cdot)$. For example, if we classify all the local patches into different classes $\{\mathbb{C}\}_{n=1}^{N}$, we can then establish the mapping of a class instead of an individual patch, *i.e.*, $x = f(y|y \in \mathbb{C}_n)$. To the best of our knowledge, recent state-of-the-art learning-based SR methods often focus on dictionary learning or network training, and combinations of traditional local descriptor methods and mapping-based SR reconstruction have not been discussed in detail. How to apply efficient local features for the fast SR scenario? To answer this question, this paper proposed an extendible SR algorithm, namely, a local patch encoding (LPE)-based method. In the proposed method, image patches are encoded into different classes according to their local distributions, and then a projection matrix is computed for each class to characterize the mapping relationship between the LR and HR patches. In the reconstruction stage of the proposed method, an input LR patch can be simply reconstructed by calculating its class-label and then multiplying the corresponding projection matrix. Furthermore, we correspondingly present a simple constraint to calculate the projection matrix for each class of local patches. Experimental results demonstrate that the proposed method can efficiently recover the HRIs.

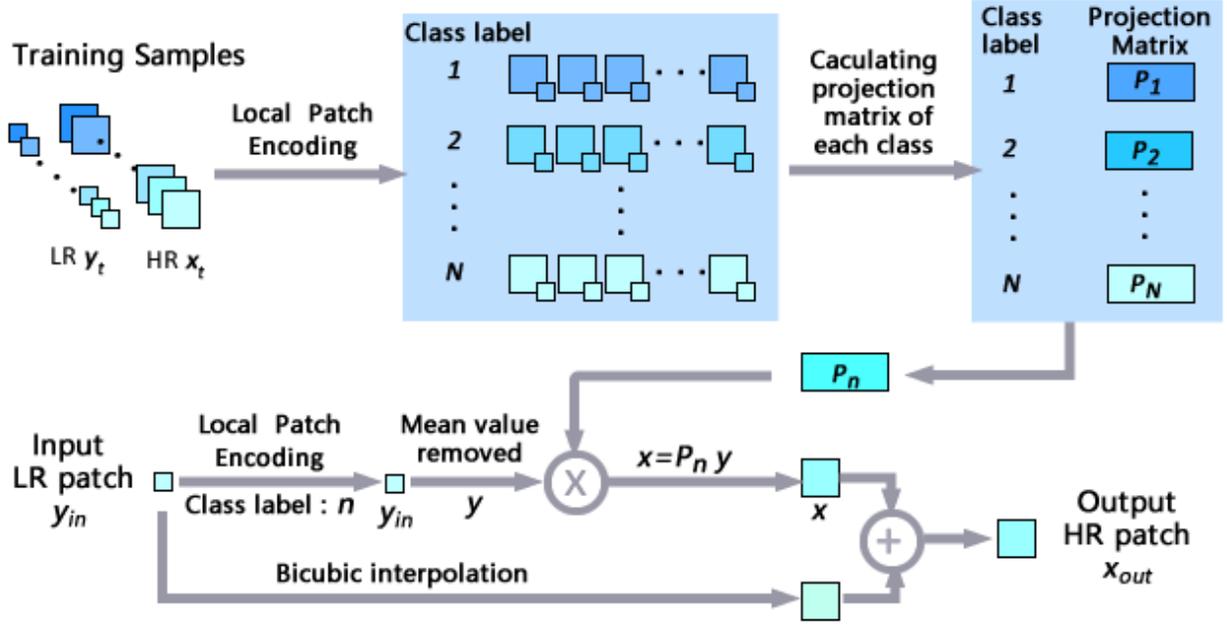

Fig. 1 The framework of the proposed method

Overall, the main technical contributions of this work are summarized as follows.

1) Traditional learning-based SR methods are often based on a learned dictionary, while this paper proposes an SR method based on traditional local descriptors. Various similar local feature extraction alogrithms can be directly applied to the SR scenario by replacing the local encoding process. Moreover, we also analyze the relationship between the proposed method and efficient ANR methods [31,32,45,47,48].

2) For classifying local patches, we propose an encoding-based method LPE to describe the local distribution. The LPE code can be directly used as the class-label, and thus no dictionary needs to be pre-learned in the proposed method. Furthermore, the discrimination capablity of LPE can be easily enhanced by directly increasing the bit-depth of the LPE code.

3) Based on the classification of local patches, we also introduce a simple way to compute the projection matrix by minimizing the total reconstruction errors of each class. Moreover, we further accelarate the reconstruction process by converting the projection matrixes to filters.

The following paragraphs are organized as follows. Section 2 presents the proposed method in detail. Section 3 gives the experimental results to testify the effectiveness of the proposed method, and Section 4 concludes the paper.

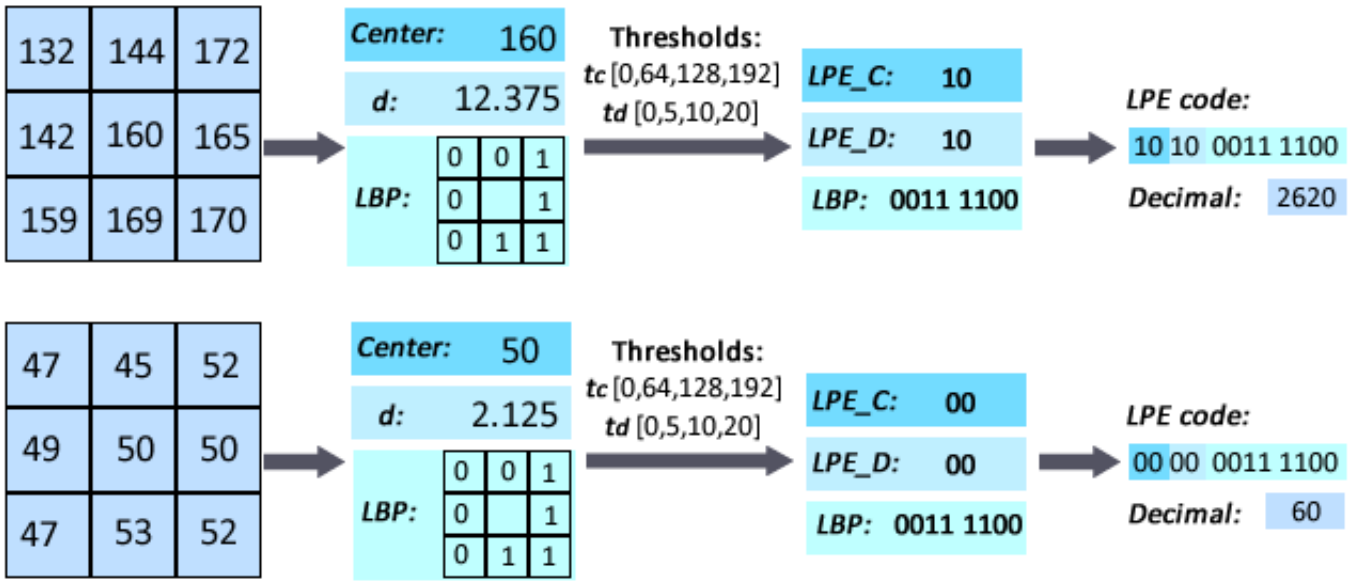

Fig.2 Illustration of the 12-bit LPE codes of two different patches

## 2. Local patch encoding-based SR method

### 2.1 Framework of the proposed method

As illustrated in Fig. 1, the proposed LPE-based SR method can be divided into two stages: a learning stage in which the local patches are classified into different classes and the projection matrix is then calculated for each class; and a reconstructing stage in which the LR input is reconstructed by means of a corresponding pre-computed projection matrix.

In the learning stage, various LR patches and corresponding HR patches are randomly selected from training images. These LR and HR patch pairs are then classified into $N$ classes by a local encoding process. As a result, the patches in the same class have a similar local pattern and grayscale distribution. Finally, a projection matrix is computed for each class to represent the mapping relationship between the LR patches and the HR patches.

In the reconstructing stage, the class-label $n$ of the input LR patch is first obtained by means of the LPE. The input patch is then multiplied with the corresponding pre-computed projection matrix $P_n$ to produce its HR patch. It should be noted that, in practice, the training patches and input patches are mean-value-removed as in many other learning-based SR methods, *e.g.*, [31,32,43]. That means traditional interpolation-based methods are used to obtain an initial patch and the learning-based processes are mainly to recover the missing high-frequency components. Hence, the reconstructed HR patch is finally combined with the bicubic-interpolated patch to obtain the HR output.

In the following, we first introduce the LPE and then present a simple constraint to calculate the projection matrix. At last, we further analyze the extendibility of the proposed method.

*2.2 Local patch encoding-based patch classification*

In the proposed method, patches with similar local distribution shall be classified into the same class. Therefore, we propose an LPE method based on a popular local descriptor of a local binary pattern (LBP) [24]. The LBP code has been widely utilized to characterize the local pattern in many texture analysis works. Usually, the LBP coding strategy is described as follows.

$$LBP = \sum_{p=1}^{N_p} s(g_p - g_c) 2^{p-1}, \qquad s(x) = \begin{cases} 1, x \geq 0 \\ 0, x < 0 \end{cases} \quad (1)$$

where $g_c$ represents the gray value of the central pixel and $g_p$ $(p = 1,2,\cdots,N_p)$ denotes the gray value of the neighboring pixels around the central pixel, and $N_p$ is the total number of neighboring pixels. However, traditional LBP merely characterizes the local structure but omits to describe the local grayscales. Hence, we propose a simple but effective local patch encoding method to distinguish different local patches via their binary structure and local gray value difference.

The gray value of the central pixel represents the grayscale level of the local patch. Hence, we quantize the central pixel value into various levels by means of an LPE code of central pixel value (LPE_C), which is defined as

$$LPE\_C = bin(Q_{c_i} | g_c \in Q_{c_i}), i = (1,2,\cdots,2^{N_c}) \quad (2)$$

where $Q_{c_i}$ $(i = 1, 2, \cdots, 2^{N_c})$ is the index of the gray value interval quantized by a series of quantization thresholds, $N_c$ denotes the bit-depth of the central pixel value encoding, and $bin(\cdot)$ denotes the binarization process. The LPE_C code thus denotes the binary index of the quantized interval covering the central gray value. For example, if $N_c$ is set as 2, the central gray value is quantized into four intervals, and the binary indexes of these four intervals are "00", "01", "10", and "11". In this paper, we utilize a series of homogeneous thresholds $t_c$ to quantize the central pixel value, and the thresholds $t_c$ is simply set as $\left[0, \frac{256}{N_c}, \frac{256}{N_c} \times 2, \cdots, \frac{256}{N_c} \times (N_c - 1)\right]$.

The mean local gray value difference is utilized to describe the magnitude of local difference, which is computed as

$$d = \frac{1}{N_p} \sum_{p=1}^{N_p} |g_p - g_c| \quad (3)$$

where $g_p$, $g_c$ and $N_p$ have been defined in Eq. (1). The mean local gray value difference $d$ can also be quantized into several levels by a set of thresholds. As a result, the LPE code of local difference (LPE_D) is defined as

$$LPE\_D = bin\left(Q_{d_j} \middle| d \in Q_{d_j}\right), j = (1,2,\cdots,2^{N_d}) \quad (4)$$

where $Q_{d_j}$ $(j = 1, 2, \cdots, 2^{N_d})$ denotes the index of quantized interval of local gray value difference, and $N_d$ is the bit-depth of the local gray value difference encoding. The LPE_D code denotes the binary index of the interval covering the mean local gray value difference $d$. In this paper, the thresholds $t_d$ used to quantize the local gray value difference are experimentally set as $[0, 5, 10, \cdots, 5 \times (N_d - 1)]$.

Finally, we obtain the LPE code of a local patch by the concatenation of the LPE_C code, the LPE_D code, and the LBP code:

$$LPE = bin(Q_{c_i}|g_c \in Q_{c_i}) \oplus bin\left(Q_{d_j}\middle| d \in Q_{d_j}\right) \oplus bin\left(\sum_{p=1}^{N_p} s(g_p - g_c) 2^{p-1}\right) \quad (5)$$

where $\oplus$ denotes the concatenation of binary codes.

In this work, we extract the LPE code from a $3 \times 3$ LR patch, and $N_p$ is thus set as 8. After the local patch encoding process, each LR patch is encoded by an LPE code with a bit-depth of $(N_c + N_d + N_p)$, and this LPE code can be regarded as the class-label of the LR patch and its corresponding HR patch. Note that the 8-bit LPE code is set as the traditional LBP code. When the bit-depth of the LPE code increases, we alternately increase the quantization levels of LPE_C and LPE_D. For example, the 12-bit LPE code is a concatenation of 2-bit LPE_C code, 2-bit LPE_D code, and 8-bit LBP code. Fig.2 illustrates a detailed calculation of the 12-bit LPE code of two local patches.

Note that there are many local patch classification methods; we utilize the simple LPE encoding in this paper for the following reasons. First, the proposed encoding process is easy to be calculated. Second, the LPE code of a local patch can be directly used as its class-label, and therefore the complex dictionary learning process or extra classifier can be omitted. For example, if a patch is encoded as "0000 0011 1100" (60), this patch can be directly classified into the 60-th class. Third, motivated by efficient LBP-variants used in local texture analysis works, the simple local encoding process can accurately characterize the local distribution by encoding the local gray value difference and local textural structure. Obviously, the proposed LPE has better discriminative capability than the traditional LBP. As shown in Fig.2, two different patches may have the same LBP code, while the proposed LPE can easily distinguish them. Lastly, the patches can be accurately classified into more classes by simply increasing the bit-depth of the LPE code, while the computation complexity of the reconstructing stage is almost unchanged.

*2.3 Derivation of projection matrix*

After the LPE process, image patches are encoded into various classes according to their local distributions. A projection matrix is then computed for each class to establish the mapping relationship between LR patches and HR patches. One basic constraint of image SR is that the upsampled LR patch should be consistent with its HR patch. Hence, we can calculate the projection matrix by minimizing the total reconstruction errors in one class. This constraint can be described as

$$\min_{P} \sum_{i=1}^{N_o} \|Py_i - x_i\|_2^2 \quad (6)$$

where $y_i$ denotes an LR patch, $x_i$ is its corresponding HR patch, $P$ is the projection matrix, and $N_o$ denotes the number of training patches in this class. It should be noted that patches $y_i$ and $x_i$ are in the form of vectors. The algebraic solution of Eq. (6) is given by

$$P = \sum_{i=1}^{N_o} x_i y_i^T (\sum_{i=1}^{N_o} y_i y_i^T)^{-1} \qquad (7)$$

where the inverse of matrix denotes the generalized inverse.

In this paper, we randomly select $N_o$ (LR and HR) patch pairs from each class to compute its projection matrix. Note that the proposed constraint is based on the assumption that local patches are accurately classified. If the patches in one class have different local distributions, it is unreasonable to reconstruct them with the same projection matrix. However, more accurate patch classification results can further improve the performance of the proposed projection matrix. Hence, this constraint is very suitable for the proposed method, which is based on accurate local patch description.

In the reconstructing stage, an LR patch $y$ with class-label $n$ can be simply reconstructed as follows

$$x = P_n y \qquad (8)$$

where $P_n$ denotes the projection matrix of the $n$-th class. In practice, $y$ is obtained by subtracting the mean value from the LR input $y_{in}$. Therefore, the final HR output $x_{out}$ is the summation of the reconstructed HR patch $x$ and the bicubic-interpolated LR input:

$$x_{out} = P_n y + H^T U y_{in} \qquad (9)$$

where $y$ denotes the mean-value-removed LR input, $U$ denotes an upsampling operator, and $H$ is a blurring operator.

The proposed algorithms for the learning stage and the reconstructing stage are summarized in **Algorithm 1** and **Algorithm 2**, respectively.

*2.4 Extendibility of the proposed SR method*

In the following, we first clarify the relationship between the ANR model and the LPE model. The extendibility of the proposed SR method is then analyzed.

In the ANR-based methods [31,32], the nearest neighbor atom of an LR input $y$ is first searched from the learned dictionary. The pre-computed projection matrix of this atom is then used to estimate the projection matrix of $y$. If we view each dictionary atom as a class-center, the ANR-based methods also roughly classify the input patches into different classes according to the distances between $y$ and each class-center. Therefore, the ANR method and the LPE method both enforce the local space partition and then establish the mapping relationships from the LR space to the HR space. But approaches to space partition by the LPE model and the ANR model differ markedly. The ANR is based on dictionary learning, while the LPE is based on local texture descriptors. The ANR methods divide the LR space according to the learned dictionary atoms, which often consist of stable edge-like patterns. The proposed LPE uniformly characterizes each kind of local distribution by means of its code, regardless of whether it is a primary pattern. In other words, the space partition of the LPE is more uniform and homogeneous. As a result, the LPE method is robust to various kinds of local textures, while the ANR methods mainly focus on the reconstruction of fine edges.

The proposed SR method can be easily extended to various methods. First, various traditional local feature extraction methods can be applied in the local patch classification process, such as gradient-based features, variants of local descriptors, and filtering-

---

**Algorithm 1.** Learning stage of the LPE based SR

---

1) **Input:** LR samples and corresponding HR samples

2) **Classification:**

   Classify LR and HR samples into total $N$ classes by means of the LPE codes computed by Eq.(5)

3) **for** each class $n$ of $N$ do

   a) Randomly select $N_o$ LR and HR samples from the $n$-th class;

   b) Removed the mean patch value from each LR and HR sample;

   c) Calculate the projection matrix $\boldsymbol{P}_n$ by means of Eq. (7).

4) **end for**

5) **Outputs**: $\boldsymbol{P}_n$ $(n = 1,2,\cdots,N)$ for each class.

---

**Algorithm 2.** Reconstruction stage of the LPE based SR

---

1) **Inputs**: LR image $Y$, magnified factor $s$, projection matrix $\boldsymbol{P}_n$ $(n = 1,2,\cdots,N)$

2) **Initialization:** set the HR image $X = 0$

3) **for** each image patch $\boldsymbol{y}_{in}$ in $Y$ do

   a) Upsample $\boldsymbol{y}_{in}$ by using bicubic interpolation with a factor of $s$;

   b) Remove the mean value from $\boldsymbol{y}_{in}$ to get $\boldsymbol{y}$;

   c) Obtain the class-label $n$ of $\boldsymbol{y}_{in}$ by computing its LPE code with Eq. (5);

   d) Reconstruct the HR output $\boldsymbol{x}_{out}$ by means of Eq. (9);

   e) Add $\boldsymbol{x}_{out}$ to the corresponding pixels in $X$.

4) **end for**

5) **Average** overlapping regions of $X$ between the adjacent patches

6) **Outputs**: HR image $X$.

based features. Second, different constraints can be utilized to compute the projection matrix for each class. For example, the ANR applied a constraint that minimizes the representation error of the dictionary atom. In the ANR-based methods, a ridge regression is used to calculate the representation of LR input feature $\boldsymbol{y}$ as follows

$$\min_{\alpha} \|\boldsymbol{y} - \boldsymbol{N}_l \alpha\|_2^2 + \lambda \|\alpha\|_2 \qquad (10)$$

where $\boldsymbol{N}_l$ denotes the neighborhood in the LR space and $\lambda$ is a weighting factor to stabilize the solution. The algebraic solution of Eq. (10) is given by

$$\alpha = (\boldsymbol{N}_l^T \boldsymbol{N}_l + \lambda \boldsymbol{I})^{-1} \boldsymbol{N}_l^T \boldsymbol{y}. \qquad (11)$$

As a result, the HR patch can be estimated by the same coefficients $\alpha$:

$$\boldsymbol{x} = \boldsymbol{N}_h \alpha = \boldsymbol{N}_h (\boldsymbol{N}_l^T \boldsymbol{N}_l + \lambda \boldsymbol{I})^{-1} \boldsymbol{N}_l^T \boldsymbol{y} \qquad (12)$$

where $\boldsymbol{N}_h$ is the HR neighborhood corresponding to $\boldsymbol{N}_l$. The projection matrix is then defined by

$$\boldsymbol{P}_2 = \boldsymbol{N}_h (\boldsymbol{N}_l^T \boldsymbol{N}_l + \lambda \boldsymbol{I})^{-1} \boldsymbol{N}_l^T. \qquad (13)$$

Note that the $l$-2 norm is utilized in both Eq. (6) and Eq. (10), so that the projection matrixes can be easily defined by means of their close-form solutions.

The computational complexity of the reconstruction process is related to the size of the projection matrix, which is the product of the input patch size and the output patch size. Compared to large projection matrixes, the filtering process can save a great deal of computational time. Hence, projection matrixes can be replaced by filters to further accelerate the reconstruction process. By converting the projection process $\boldsymbol{Py}_i$ to filtering process $\boldsymbol{P}_3 * \boldsymbol{y}_i$, the former constraint Eq. (6) becomes

$$\min_{\boldsymbol{P}_3} \sum_{i=1}^{N_o} \|\boldsymbol{P}_3 * \boldsymbol{y}_i - \boldsymbol{x}_i\|_2^2 \qquad (14)$$

Instead of introducing the solution with Fast Fourier Transformation (FFT), we solve this optimizer by transforming the reconstruction of the image patch to that of a single central pixel, as in [25]. Thus, in Eq. (14), $\boldsymbol{y}_i$ denotes the vector form of a local patch around a central pixel, and $\boldsymbol{x}_i$ becomes the HR ground truth value of this central pixel. Then, the filtered result of the central pixel can be represented by the product of the filter vector $\boldsymbol{P}_3$ and local patch vector $\boldsymbol{y}_i$. In order to compute the filter of each class, we collect the training patches and corresponding HR pixel values to form two matrixes for every class, *i.e.*, $\boldsymbol{A} = \sum_{i=1}^{N_I}(y_{i1}, y_{i2}, \cdots, y_{iM})^T$, and $\boldsymbol{B} = \sum_{i=1}^{N_I}(x_{i1}, x_{i2}, \cdots, x_{iM})^T$. $y_{i1}, y_{i2}, \cdots, y_{iM}$ denote that the total $M$ local patches of the $i$-th image, $x_{i1}, x_{i2}, \cdots, x_{iM}$, are corresponding HR values of the central pixels, and $N_I$ denotes the total number of training images. The filter of one class is then calculated by simply solving the following least-square problem

$$\min_{\boldsymbol{P}_3} \|\boldsymbol{A}\boldsymbol{P}_3 - \boldsymbol{B}\|_2^2 \qquad (15)$$

The benefit and drawback of the filter $P_3$ are both prominent. The filtering process is faster than the projection reconstruction. In this paper, we adopt a 3 × 3 filter to achieve the least computational complexity. On the other hand, the optimized filtering computed by Eq. (15) is indeed the reconstruction of the central pixel. However, the reconstruction of a single pixel is less robust than the reconstruction of an entire local patch.

In this paper, we use the projection matrix $P_1$ computed by Eq. (7), the projection matrix $P_2$ computed with the traditional ridge regression, and the filter $P_3$. The LPE-based SR methods with $P_1$, $P_2$, and $P_3$ are denoted by "LPE*xx*bits_P$_1$", "LPE*xx*bits_P$_2$", and "LPE*xx*bits_P$_3$", respectively, where "*xx*bits" denotes the bit-depth of the LPE code.

## 3. Experiments

*3.1 Testing image sets*

We test the proposed method on three testing image sets, *i.e.*, "Set5", "Set14", and "B100". "Set 5" and "Set14"[1] [46] contain 5 and 14 commonly used images for SR evaluation, respectively. "B100" [31] consists of 100 testing images selected from the Berkeley segmentation dataset (BSD) [22]. For color images, it is first converted from RGB to YCbCr. Various SR methods are then applied only on the Y (intensity) component, and bicubic interpolation is used for the other components. In our experiments, the input LRIs are obtained by downsampling the original HRIs with bicubic interpolation and the LRIs are then upsampled to their original size with different methods. Note that the ASDS [6] method first filtered and then downsampled the HRIs to obtain the LRIs, which differs slightly with other methods. The upsampling factors in our experiment are set as 2, 3, and 4, respectively.

*3.2 Compared methods and implementation details*

In this paper, the proposed method is compared with many typical learning-based SR methods, such as the LLE [2], the ScSR [43], the Zeyde's method [46], the ASDS [6], the ANR [31], and the A+ [32]. In addition, we also use some recent state-of-the-art convolutional neural network (CNN)-based SR methods as further comparison, such as the SRCNN [4] and the VDSR [16].

For calculating the LPE codes, we select various sets of thresholds to quantize the central gray value and the mean local gray value difference. We experimentally select two sets of thresholds as in section 2.2. The sizes of local patches are set as 6, 9, and 12 for 2 ×, 3 ×, and 4 × magnification, respectively. For calculating the projection matrix with Eq. (13), we use the same factor $\lambda$ as in [31].

We used the same training set proposed by Yang *et al.* [43], which is also used in many other methods. Fig. 3 illustrates the average PSNR values on "Set5" and "Set14" with different numbers of patches that are used to calculate the projection matrix in each class. We can find that more samples always obtain better results. Hence, we randomly select 2048 LR and HR patch pairs from each class. Note that the training set only contains 91 images and thus the total numbers of patches may be less than 2048 in some classes. On this condition, we used all the patches in one class to compute its projection matrix.

---

[1] Note that the image "*Bridge*" of "Set 14" is a zero-padded 6-bit image, and it is unreasonable to use a 6-bit image to verify the performance of the SR methods trained with 8-bit images. Hence, in our experiment, we have removed the low bit-depth image from the original "Set14".

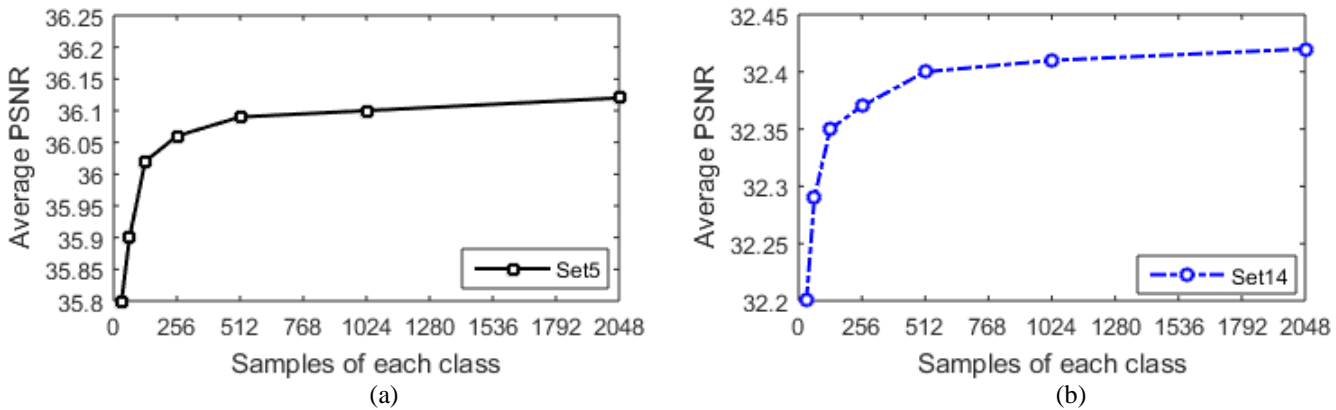

Fig. 3 Average PSNR (dB) of LPE_P$_1$ (2✕ magnification) with different numbers of selected samples in each class, (a) on image set "Set5", (b) on image set "Set14".

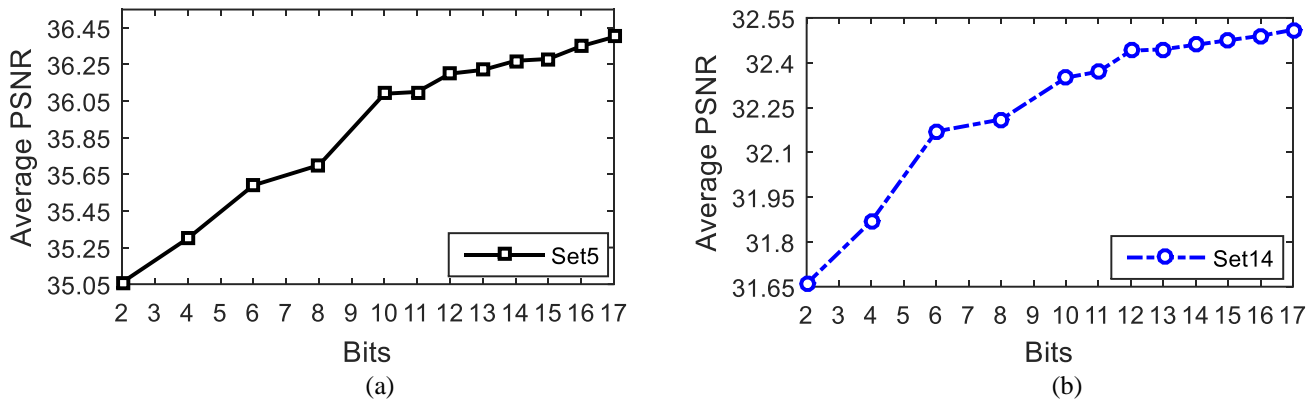

Fig.4 Average PSNR (dB) of LPE_P$_1$ (2✕ magnification) with different bits depth of the LPE code, (a) on image set "Set5", (b) on image set "Set14".

As illustrated in Fig.4, we can find that better performance on "Set5" and "Set14" can be achieved when the bit-depth of the LPE code increases, since the characterization of the local patch becomes more accurate when the bit-depth of the LPE code increases. In Fig.4, we also test the low-bits, which are less than 8. Because the LBP code costs 8-bits, the LBP code is thus not used in these LPE codes which are less than 8-bits. For instance, the extreme 2-bit LPE only consists of 1-bit LPE_C code and another 1-bit LPE_D code. Although the computation cost of the LPE process increases little when the bit-depth of the LPE code increases, much more projection matrixes have to be pre-computed. For example, $2^{12}$ (4096) projection matrixes are computed for 12-bit LPE code, and $2^{17}$ (131072) projection matrixes are required to be calculated for 17-bit LPE code. We thus mainly utilize the 12-bit LPE codes in the experiment to balance the quality of SR results and the number of pre-computed matrixes. By increasing the bit-depth of the LPE code, the performance of the proposed method can be further improved. Note that it is suggested that the number of training images also be increased to ensure that there are enough LR and HR samples in each class when the bit-depth of the LPE code increases.

*3.3 Experimental results*

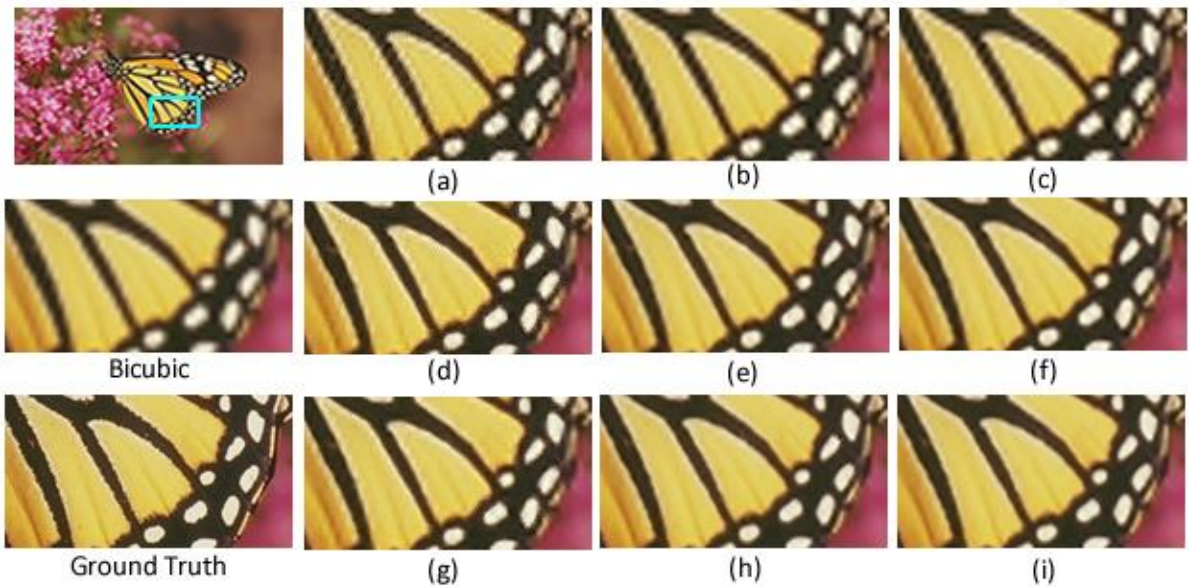

Fig.5. SR results (3×) of the proposed method with different LPE bit-depths and projection matrixes, (a) the LPE2bits_$P_3$, (b) the LPE2bits_$P_2$, (c) the LPE2bits_$P_1$, (d) the LPE10bits_$P_3$, (e) the LPE10bits_$P_2$, (f) the LPE10bits_$P_1$, (g) the LPE12bits_$P_3$, (h) the LPE12bits_$P_2$, (i) the LPE12bits_$P_1$.

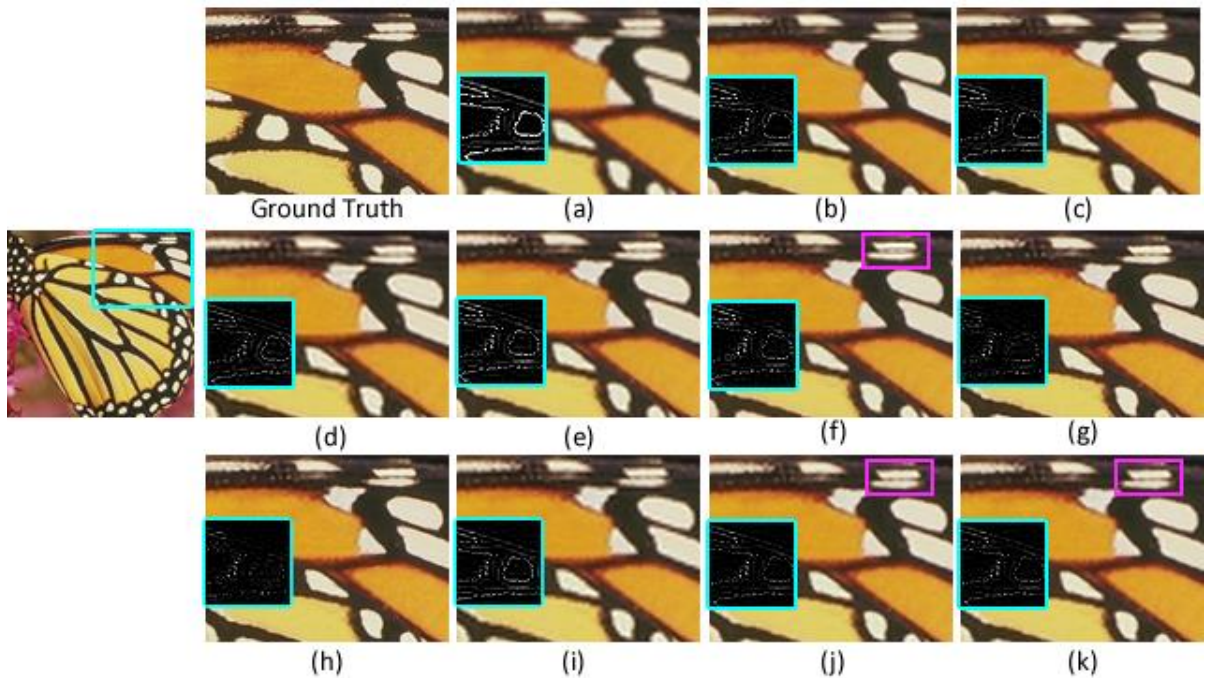

Fig.6. SR results of "monarch" image with different methods (2×), (a) bicubic, (b) the LLE [2], (c) the ScSR [43], (d) the ASDS [6], (e) the ANR [31], (f) the A+ [32], (g) the SRCNN [4], (h) the VDSR [16], (i) the LPE12bits_ $P_3$, (j) the LPE12bits_$P_2$, (k) the LPE12bits_$P_1$. The selected area in the blue square show the residual map between each result and ground truth.

Fig. 5 illustrates the SR performances of the proposed method with different encoding bits and projection matrixes. By comparing the results with various LPE bits, we find that increasing the bits-depth of the LPE code can obtain finer subjective quality. Note that the 2-bit LPE performs much worse than 10-bit and 12-bit LPE because the 2-bit LPE merely divides local patches into four classes. A comparison of the fast filters $P_3$ and projection matrixes $P_1/P_2$ shows that LPE_$P_1$/$P_2$ is more robust than LPE_$P_3$. As mentioned previously, the reconstruction of a single pixel is not as robust as the mapping of an entire patch; the SR results of LPE_ $P_3$ thus may contain some unnatural noises caused by incorrect reconstruction of some pixels.

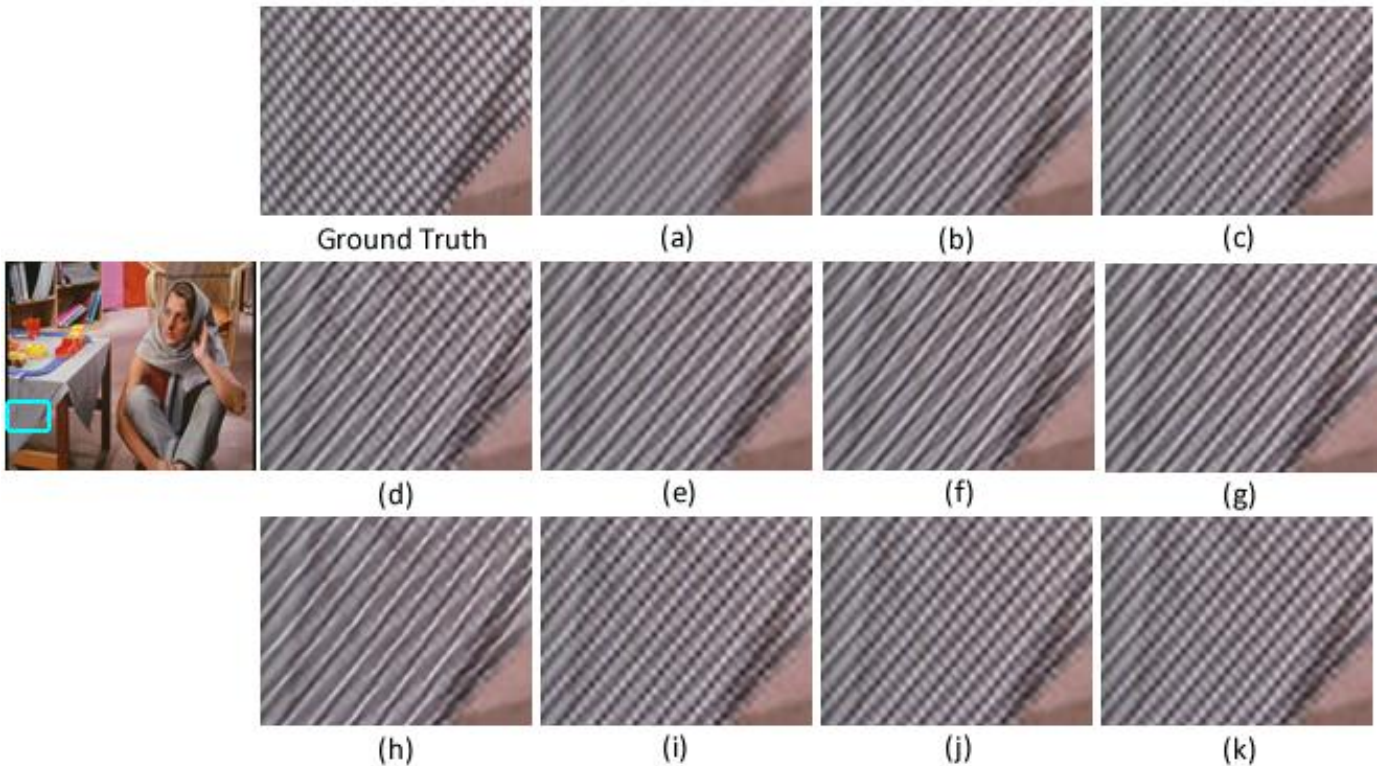

Fig.7. SR results of "barbara" image with different methods (3×), (a) bicubic, (b) the LLE [2], (c) the ScSR [43], (d) the ASDS [6], (e) the ANR [31], (f) the A+ [32], (g) the SRCNN [4], (h) the VDSR [16], (i) the LPE12bits_ $P_3$, (j) the LPE12bits_$P_2$, (k) the LPE12bits_$P_1$.

Fig. 6 shows the SR results of the "monarch" image using different methods for 2× magnification. The following observations are derived from Fig.6. First, the bicubic interpolation produces blurry edges, and these learning-based methods can recover sharp edges. Second, by comparing the tiny edges that are marked in the red rectangle, the proposed LPE methods can reproduce better details than the A+. Third, by comparing the LPE methods with different projection matrixes, LPE_$P_1$ and LPE_$P_2$ achieve similar performance. The fast LPE_$P_3$ can recover sharp edges, but the details of the LPE_$P_3$ result are worse than those of LPE_$P_1$/$P_2$. Fourth, the SRCNN and the VDSR can recover obviously sharp and clear results, and the VDSR reproduces the sharpest edges among these methods. Overall, LPE_$P_1$/$P_2$ can achieve comparable visual quality to these state-of-the-art methods. Furthermore, the residual component between each SR result and the ground truth HRI is also illustrated in the blue square. By comparing these residual maps, it can be found that the LPE_$P_1$/$P_2$ and the CNN-based methods can produce results exhibiting less difference with the ground truth than the ANR methods.

Fig. 7 illustrates the 3× magnification results of a texture area on the "barbara" image. First, we compare the proposed method with the ANR-based methods. As mentioned before, the ANR methods are based on dictionary learning, while the LPE method is based on the local descriptor method. In the dictionary learning process, the learned atoms often consist of stable local structures, such as local edge-like patterns. Unfortunately, the local texture patterns, which contain abundant high-frequency details, are often omitted in these atoms. Compared with learned/clustered atoms, the local encoding process is widely utilized to characterize the local texture in many texture analysis algorithms, and can extract non-edge-like patterns much better. As illustrated in Fig.7, the cross lines have different widths, and the thinner lines become more blurry during downsampling. The ANR can obviously sharpen the main edges but suppress the tiny edges. As a result, the thicker lines are over-sharpened and the thinner lines are further blurred. In the A+ result, the thick lines become sharper, and the tiny structures have nearly disappeared. Although the ANR methods can reproduce sharp edges, the reconstructed texture may be different from the original image. Similarly, the VDSR and the SRCNN also seriously over-strengthen the main edges but damage the textural details. Compared to these state-of-the-art methods, the local encoding process can faithfully describe the local distribution, regardless of whether it is a primary pattern. Hence, the SR results of the LPE method are more consistent with the original texture.

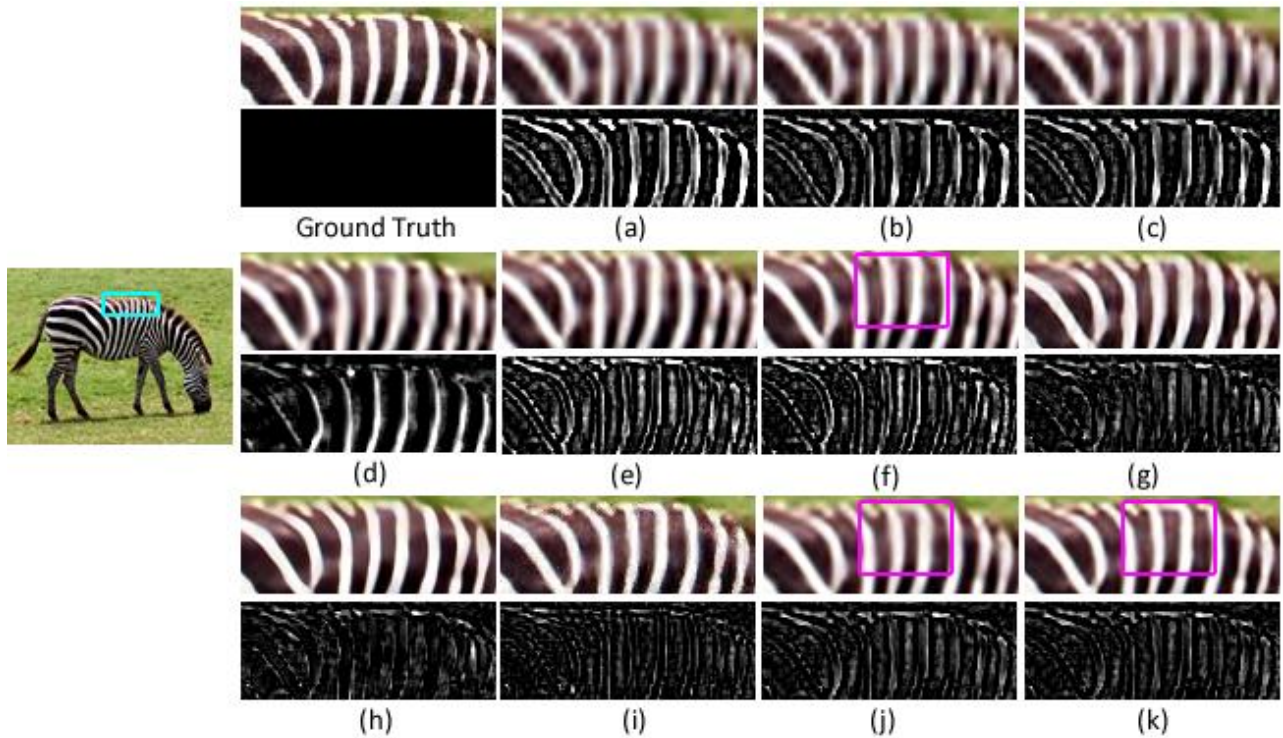

Fig.8. SR results of "zebra" image with different methods (4×), (a) bicubic, (b) the LLE [2], (c) the ScSR [43], (d) the ASDS [6], (e) the ANR [31], (f) the A+ [32], (g) the SRCNN [4], (h) the VDSR [16], (i) the LPE12bits_ $P_3$, (j) the LPE12bits_$P_2$, (k) the LPE12bits_$P_1$. The residual map between each SR result and ground truth is also given.

Some 4× magnification results of the "zebra" image are shown in Fig.8. By comparing the boundaries of streaks, we obtain findings similar to those in Fig.6. First, by comparing the edge details marked in the red rectangle, the LPE-based methods can obtain sharper edges than the A+ for 4× magnification. Second, the LPE_$P_3$ may produce some slight noise around the reconstructed edges. Third, the VDSR still recovers the sharpest edges. Finally, by comparing the residual maps of these methods, the LPE methods still exhibit fewer differences with the HRI than the ANR methods.

Table 1. Average PSNR (dB), SSIM, and IFC of different methods on image set "Set5"

| | 2× | | | 3× | | | 4× | | |
|---|---|---|---|---|---|---|---|---|---|
| | **PSNR** | **SSIM** | **IFC** | **PSNR** | **SSIM** | **IFC** | **PSNR** | **SSIM** | **IFC** |
| *Bicubic* | 33.66 | 0.9383 | 5.72 | 30.40 | 0.8804 | 3.44 | 28.44 | 0.8250 | 2.29 |
| *LLE [2]* | 35.57 | 0.9543 | 6.08 | 31.73 | 0.8836 | 4.24 | 29.64 | 0.8510 | 2.93 |
| *ScSR [43]* | 35.38 | 0.9564 | 7.00 | 31.23 | 0.9068 | 4.15 | 29.43 | 0.8551 | 2.69 |
| *ASDS [6]* | 34.85 | 0.9544 | 6.81 | 31.02 | 0.9003 | 4.31 | 29.54 | 0.8497 | 2.75 |
| *Zeyde's [46]* | 35.64 | 0.9559 | 6.87 | 31.79 | 0.9032 | 4.37 | 29.69 | 0.8533 | 2.70 |
| *ANR [31]* | 35.82 | 0.9568 | 7.30 | 31.84 | 0.9072 | 4.39 | 29.68 | 0.8558 | 3.00 |
| *LPE2bits_P$_2$* | 35.07 | 0.9507 | 7.05 | 31.27 | 0.8955 | 4.09 | 29.12 | 0.8420 | 2.69 |
| *LPE2bits_P$_1$* | 35.06 | 0.9509 | 7.07 | 31.28 | 0.8957 | 4.09 | 29.16 | 0.8423 | 2.70 |
| *LPE10bits_P$_3$* | 34.77 | 0.9420 | 6.01 | 30.96 | 0.8823 | 3.99 | 28.76 | 0.8331 | 2.54 |
| *LPE10bits_P$_2$* | 36.12 | 0.9573 | 7.31 | 32.22 | 0.9132 | 4.44 | 30.03 | 0.8729 | 3.11 |
| *LPE10bits_P$_1$* | 36.10 | 0.9572 | 7.30 | 32.27 | 0.9125 | 4.40 | 30.09 | 0.8720 | 3.09 |
| *LPE12bits_P$_3$* | 35.02 | 0.9445 | 6.09 | 31.19 | 0.8907 | 4.06 | 29.02 | 0.8381 | 2.66 |
| *LPE12bits_P$_2$* | 36.14 | **0.9585** | 7.37 | 32.33 | **0.9165** | 4.49 | 30.19 | 0.8764 | **3.19** |
| *LPE12bits_P$_1$* | **36.19** | 0.9582 | **7.39** | **32.41** | 0.9161 | **4.57** | **30.22** | **0.8773** | 3.13 |

Table 2. Average PSNR (dB), SSIM, and IFC of different methods on image set "Set14"

| | 2× | | | 3× | | | 4× | | |
|---|---|---|---|---|---|---|---|---|---|
| | **PSNR** | **SSIM** | **IFC** | **PSNR** | **SSIM** | **IFC** | **PSNR** | **SSIM** | **IFC** |
| *Bicubic* | 30.36 | 0.9417 | 5.83 | 27.67 | 0.8596 | 3.41 | 26.12 | 0.7857 | 2.27 |
| *LLE [2]* | 31.91 | 0.9587 | 6.08 | 28.74 | 0.8836 | 3.89 | 26.95 | 0.8137 | 2.21 |
| *ScSR [43]* | 31.21 | 0.9620 | 6.22 | 28.01 | 0.8882 | 4.04 | 26.57 | 0.8183 | 2.65 |
| *ASDS [6]* | 31.15 | 0.9627 | 6.61 | 27.91 | 0.8938 | 4.11 | 26.94 | 0.8190 | 2.35 |
| *Zeyde's [46]* | 31.96 | 0.9589 | 6.25 | 28.80 | 0.8841 | 4.02 | 26.99 | 0.8159 | 2.67 |
| *ANR [312]* | 31.95 | 0.9626 | 6.36 | 28.80 | 0.8890 | 3.67 | 27.00 | 0.8194 | 2.48 |
| *A+ [32]* | 32.39 | 0.9641 | 6.54 | 29.12 | 0.8940 | 4.04 | 27.34 | 0.8294 | 2.62 |
| *LPE2bits_P$_2$* | 31.65 | 0.9614 | 6.40 | 28.44 | 0.8932 | 3.95 | 26.61 | 0.8348 | 2.49 |
| *LPE2bits_P$_1$* | 31.66 | 0.9617 | 6.44 | 28.44 | 0.8934 | 3.97 | 26.63 | 0.8347 | 2.44 |
| *LPE10bits_P$_2$* | 32.35 | 0.9677 | 6.86 | 29.11 | 0.9013 | 4.09 | 27.33 | 0.8398 | 2.69 |
| *LPE10bits_P$_1$* | 32.39 | 0.9674 | 6.82 | 29.15 | 0.9016 | 4.13 | 27.35 | 0.8379 | 2.71 |
| *LPE12bits_P$_3$* | 31.08 | 0.9498 | 6.02 | 27.89 | 0.8839 | 3.83 | 26.44 | 0.8109 | 2.32 |
| *LPE12bits_P$_2$* | 32.44 | 0.9688 | 7.27 | 29.19 | 0.9088 | 4.22 | 27.40 | **0.8603** | 2.87 |
| *LPE12bits_P$_1$* | **32.47** | **0.9692** | **7.27** | **29.21** | **0.9093** | **4.23** | **27.42** | 0.8596 | 2.84 |

Table 3. Average PSNR (dB), SSIM, and IFC of different methods on image set "B100"

| | 2× | | | 3× | | | 4× | | |
|---|---|---|---|---|---|---|---|---|---|
| | **PSNR** | **SSIM** | **IFC** | **PSNR** | **SSIM** | **IFC** | **PSNR** | **SSIM** | **IFC** |
| *Bicubic* | 29.35 | 0.8334 | 5.85 | 27.17 | 0.7361 | 3.47 | 25.95 | 0.6671 | 2.29 |
| *LLE [2]* | 30.40 | 0.8674 | 6.12 | 27.84 | 0.7687 | 3.95 | 26.47 | 0.6937 | 2.74 |
| *ScSR [43]* | 30.32 | 0.8709 | 6.24 | 27.74 | 0.7719 | 4.22 | 26.33 | 0.6997 | 2.85 |
| *ASDS [6]* | 30.19 | 0.8712 | 6.72 | 27.65 | 0.7735 | 4.24 | 26.45 | 0.7003 | 2.97 |
| *Zeyde's [46]* | 30.40 | 0.8682 | 6.32 | 27.87 | 0.7693 | 4.18 | 26.51 | 0.6963 | 2.76 |
| *ANR [31]* | 30.50 | 0.8706 | 6.59 | 27.90 | 0.7724 | 4.16 | 26.52 | 0.6991 | 2.67 |
| *A+ [32]* | 30.76 | 0.8762 | 6.64 | 28.18 | 0.7764 | 4.19 | 26.76 | 0.7062 | 2.72 |
| *SRCNN [4]* | 31.36 | 0.8879 | **7.24** | 28.41 | 0.7863 | -- | 26.90 | 0.7101 | 2.41 |
| *VDSR [16]* | **31.90** | **0.8960** | 7.17 | **28.82** | 0.7976 | -- | **27.29** | 0.7251 | 2.63 |
| *LPE10bits_P2* | 30.74 | 0.8796 | 7.04 | 28.18 | 0.7885 | 4.20 | 26.80 | 0.7148 | 2.96 |
| *LPE10bits_P1* | 30.78 | 0.8792 | 7.06 | 28.20 | 0.7863 | 4.22 | 26.82 | 0.7151 | 2.90 |
| *LPE12bits_P3* | 30.42 | 0.8711 | 6.67 | 27.82 | 0.7720 | 4.25 | 26.44 | 0.6997 | 2.84 |
| *LPE12bits_P2* | 30.97 | 0.8825 | 7.12 | 28.26 | 0.7937 | 4.34 | 26.87 | 0.7223 | 3.12 |
| *LPE12bits_P1* | 30.99 | 0.8827 | 7.14 | 28.27 | 0.7946 | 4.32 | 26.94 | 0.7223 | 3.06 |
| *LPE17bits_P2* | 31.26 | 0.8924 | 7.17 | 28.38 | **0.7977** | **4.38** | 26.99 | 0.7293 | **3.18** |
| *LPE17bits_P1* | 31.24 | 0.8927 | 7.19 | 28.39 | 0.7969 | 4.36 | 27.06 | **0.7294** | 3.11 |

The objective quality of these methods is compared on three image datasets of "Set5", "Set14", and "B100". Two commonly used quantitative evaluation metrics PSNR and SSIM are also adopted in this paper. However, the PSNR may not always reflect SR subjective quality well [43]. In [42], Yang *et al.* observed that the information fidelity criterion (IFC) [28] index has the highest correlation with perceptual scores for SR evaluation. Hence, the IFC index is also used in our experiment to verify the subjective quality of the SR results.

Table 1 lists the objective assessment of 2×, 3×, and 4× magnification on "Set5", from which we can get the following findings. First, the LPE method performs better when the bit-depth of LPE increases. Second, the LPE_$P_1$ and the LPE_$P_2$ obtain similar results. This demonstrates that different projection matrixes can be suitable for the proposed LPE method. The objective quality of the LPE_$P_3$ is much lower than the LPE_$P_1$/$P_2$, which also verifies that the reconstruction of a single pixel is less robust than the mapping of a local patch. Third, although the 2-bit LPE merely contains four projection matrixes or filters, it still performs much better than bicubic interpolation. Finally, LPE_$P_1$/$P_2$ can achieve higher PSNR, SSIM, and IFC values than some traditional learning-based SR methods.

Table 2 lists the results on "Set14", and similar observations can be obtained. First, LPE_$P_1$/$P_2$ still outperforms the fast $P_3$ mode. Second, 12-bit LPE achieves higher objective assessment values than 10-bit and 2-bit LPE. Third, the LPE12bits_$P_1$/$P_2$ can obtain better results than the ANR-based methods. This also indicates the effectiveness of the proposed LPE process.

Table 3 lists the results of different methods on the largest image set "B100". Two state-of-the-art CNN-based methods of the SRCNN [4] and the VDSR [16] are added as comparisons, and their quality assessment values are listed as reported in [4,16,17]. We can obtain the following findings from this table. First, 12-bit LPE_$P_1$/$P_2$ still performs better than the ANR-based methods. Second, CNN-based methods can achieve high PSNR values by utilizing deep network and mean square error (MSE) loss. The VDSR achieves much higher PSNR values than other methods. It also should be noted that the VDSR is trained with Yang's 91 images [43] and 200 images from the BSD dataset [22], the SRCNN is trained on the detection training set of ImageNet, and the non-CNN-based methods are all trained with Yang's 91 images. A comparison of the SSIM values indicates that the VDSR performs the best on 2× magnification, and the 17-bit LPE method achieves the highest value on 4 × SR. Third, by increasing the bit-depth to 17, the performance of the proposed method can be further improved. But note that 17-bit LPE also requires more pre-computed projection matrixes, and the CNN-based method can also be improved by utilizing a larger training set and deeper network. By comparing both the subjective and objective results between the LPE and the VDSR, it can be found that the LPE method can faithfully recovered better textural details with low computational complexity. Furthermore, the LPE is a flexible local classification based SR framework, and can obtain fine objective results. The VDSR can achieve high PSNR values and reproduce sharper edges. Overall, the proposed LPE is still a high-performance and high-preserving non-deep-learning SR algorithm.

*3.4 Discussion of the computational complexity*

Table 4. Reconstruction time of different SR method (3 × magnification).

| Method | Time (s) |
|---|---|
| *ScSR [43]* | 267.10 |
| *ASDS [6]* | 866.82 |
| *SRCNN (CPU) [4]* | 10.59 |
| *ANR [31]* | 1.37 |
| *LPE12bits_P$_1$* | 0.96 |
| *LPE12bits_P$_2$* | 0.97 |
| *LPE12bits_P$_3$* | 0.28 |

The computational time of the SR methods depends on the image size. Table 4 lists the 3 × reconstruction time for the image "*Lenna*" (512 × 512) of some learning-based methods using MATLAB on an Intel Core i5-3317U laptop PC. It can be found that the ANR and the LPE-based methods are faster than the traditional learning-based methods of the ScSR and the ASDS. The LPE with $P_1/P_2$ is faster than the ANR, and the accelerated $P_3$ mode can be much faster. Note that the SRCNN is also implemented in CPU mode as are other methods; in practice, it can run much faster with the GPU and accelerated version [5].

In the following, we analyze the computational complexity of the LPE methods and the ANR methods. Reconstructions with the projection matrixes of the ANR and the LPE_P$_1$/P$_2$ have the same complexity. Given a local patch of size $m \times m$ and a reconstructed patch of size $n \times n$, the size of the projection matrix is thus $n^2m^2$, and the complexity of various projection reconstructions are $O(n^2m^2m^2)$. Hence, the main difference between the complexities of the LPE_P$_1$/P$_2$ and the ANR methods is the selection of projection matrix. For the ANR methods, the nearest atom of a local patch is first searched. Suppose that the size of the learned dictionary is $dm^2$, and the complexity of atom selection is $O(dm^2m^2)$. Note that this selection is computed for each patch; given an image of size $M \times N$, the total complexity of selection process becomes $O(MNdm^2m^2)$. For the LPE, the encoding process is executed one time for an entire image. The complexity is only $O((M-1)(N-1)T)$, where $T$ is the number of comparisons between matrixes. For example, the values of $T$ are 2, 8, and 14 for 2-bit, 8-bit, and 12-bit LPE, respectively. On the 512 × 512 image "*Lenna*", the atom selection of the ANR costs a total of 480ms, while the LPE merely costs around 90ms. For this reason, the LPE process is faster than the ANR methods. By further accelerating the proposed method with the $P_3$ (3 × 3 filter) mode, the complexity of the filtering process is merely $O(3^2 log3^2)$, which is many times less the complexity $O(n^2m^2m^2)$ of the projection reconstruction. Note that the LPE_P$_3$ runs several times faster than the LPE_P$_1$/P$_2$ on image "*Lenna*". Theoretically, however, the LPE_P$_3$ can be much faster, and these LPE methods still have much room for acceleration, such as implementation with C or CUDA.

## 4. Conclusions

In this paper, we proposed an effective and extendible image super-resolution method without a pre-learned dictionary, namely, the local patch encoding (LPE)-based method. In the proposed method, local patches are encoded into different classes by means of the LPE. The projection matrix of each class is then computed by utilizing a simple constraint. In the reconstructing stage, a patch can be simply reconstructed by computing its LPE code and then multiplying the LR patch with the corresponding projection matrix. Experimental results on several image sets demonstrate that the proposed method can achieve comparable results with some state-of-the-art methods. Furthermore, the proposed method is easily extended, and many non-dictionary-learning local pattern classification methods can also be directly introduced to the SR scenario by means of the proposed method.

## Acknowledgements

This work was partly supported by the grant of National Science Foundation of China 61402018, 61673157, the fundamental research funds for the Central Universities of China JZ2017HGTB0189, the grant of National Science Foundation of China 61602146, 61672063, 61702154, China 863 project of 2015AA015905, and Shenzhen Peacock Plan JCYJ 201605061722227337.

The authors would like to sincerely thank the anonymous reviewers, and Dr. Shujie Li for helpful advice during this work. The authors also sincerely thank W. S. Dong, R. Zeyde, R. Timofte, and C. Dong for sharing the source codes of the ASDS, the Zeyde's, the ANR, and the SRCNN methods.